\documentclass[letterpaper, 10 pt, conference]{ieeeconf}

\usepackage{amsmath,amsfonts,bm}

\def\1{\bm{1}}

\def\vb{{\bm{b}}}

\DeclareMathAlphabet{\mathsfit}{\encodingdefault}{\sfdefault}{m}{sl}
\SetMathAlphabet{\mathsfit}{bold}{\encodingdefault}{\sfdefault}{bx}{n}

\usepackage{hyperref}
\usepackage{url}
\usepackage[utf8]{inputenc} % allow utf-8 input
\usepackage[T1]{fontenc}    % use 8-bit T1 fonts
\usepackage{hyperref}       % hyperlinks
\usepackage{url}            % simple URL typesetting
\usepackage{booktabs}       % professional-quality tables
\usepackage{amsfonts}       % blackboard math symbols
\usepackage{nicefrac}       % compact symbols for 1/2, etc.
\usepackage{microtype}      % microtypography
\usepackage{xcolor}         % colors
\usepackage{amsmath}
\usepackage{amssymb}
\usepackage{mathtools}
\usepackage{comment}
\usepackage{algorithmic}
\usepackage{algorithm}
\usepackage{physics}
\usepackage[doi=false,url=false, backend=bibtex,natbib=true,sorting=none]{biblatex}
\usepackage{fancyhdr}
\usepackage{ifthen}
\title{Value Explicit Pretraining for Learning Transferable Representations}
\IEEEoverridecommandlockouts
\author{
  Kiran Lekkala, Henghui Bao, Sumedh A. Sontakke, Erdem Bıyık, Laurent Itti%
\thanks{Manuscript received: Month Day, Year; Revised: Month Day, Year; Accepted: January 11, 2026. This paper was recommended for publication by Editor Aleksandra Faust upon evaluation of the Associate Editor and Reviewers comments.}
\thanks{This work was supported by the National Eye Institute (grant R61EY037527), the National Science Foundation (award 2318101), and a research grant from Sandia National Laboratories. The authors affirm that the views expressed herein are solely their own, and do not represent the views of the United States government or any agency thereof.}
\thanks{All authors are with Thomas Lord Department of Computer Science at the University of Southern California. Correspondence: kiran.klekk@gmail.com. Digital Object Identifier (DOI): see top of this page.}%
}

\usepackage{titlesec}
\titlespacing*{\section}{0pt}{.8ex plus .4ex minus .4ex}{.5pc}
\titlespacing*{\subsection}{0pt}{.4ex plus .2ex minus .2ex}{.3pc}

\expandafter\def\expandafter\normalsize\expandafter{%
    \normalsize%
    \setlength\abovedisplayskip{3pt}%
    \setlength\belowdisplayskip{3pt}%
    \setlength\abovedisplayshortskip{2pt}%
    \setlength\belowdisplayshortskip{2pt}%
    }

\fancypagestyle{plain}{%
  \fancyhf{}%
  \fancyhead[L]{\scriptsize IEEE ROBOTICS AND AUTOMATION LETTERS. PREPRINT VERSION. ACCEPTED JANUARY, 2026}%
  \fancyhead[R]{\thepage}%
}
\renewcommand{\headrulewidth}{0pt}
\bibliography{paper}
\begin{document}
\maketitle
\thispagestyle{plain}

\begin{abstract}
Understanding visual inputs for a given task amidst varied changes is a key challenge posed by visual reinforcement learning agents. We propose \textit{Value Explicit Pretraining} (VEP), a method that learns generalizable representations for transfer reinforcement learning. VEP enables efficient learning of new tasks that share similar objectives as previously learned tasks, by learning an encoder that trains representations to be invariant to changes in environment dynamics and appearance. To pretrain the encoder with \textit{suboptimal unlabeled demonstration data} (sequence of observations and sparse reward signals), we use a self-supervised contrastive loss that enables the model to relate states across different tasks based on the Monte Carlo value estimate that is reflective of task progress, resulting in temporally smooth representations that capture the objective of the task. A major difference between our method and the existing approaches is the use of suboptimal unlabeled data that do not always solve the task. Experiments on Ant locomotion, a realistic navigation simulator and the Atari benchmark show that VEP outperforms current SoTA pretraining methods on the ability to generalize to unseen tasks. VEP achieves up to $2\times$ improvement in rewards, and up to $3\times$ improvement in sample efficiency. For videos of VEP policies, visit our \href{https://sites.google.com/view/value-explicit-pretraining/}{website}.
%This method could be used to transfer learned policies/skills to unseen related tasks.
\end{abstract}

\section{Introduction}
While performing everyday tasks, humans have an innate ability to appropriately extract information from what they perceive. This is often regardless of various changes related to the appearance and the dynamics of the tasks. This ability stems from understanding the objective of the tasks. While this ability is natural to humans, we need to equip robots with generalizable representations of their visual observations to achieve the same advantages. 

Unfortunately, learning generalizable representations for control is still an open problem in visual sequential decision-making. Typically in such representation learning works, an encoder $f_\phi$ is learned using a large offline dataset via a predetermined objective function. Subsequently, $f_\phi$ is used for control by mapping high-dimensional visual observations from the environment $\vb{o}_{:t}$ into a lower-dimensional latent representation $\vb{z}_t$. The representation $\vb{z}_t$ is fed into a policy $\pi(\cdot\mid\vb{z}_t)$ to generate an action $\vb{a}_t$ to solve a task. The key question in visual representation learning is: \emph{what should the learned $f_\phi$ be?} 

The challenge in learning $f_\phi$ mainly lies in discovering the correct \textit{inductive biases} that yield representations that can be used to learn a variety of downstream tasks in a sample efficient manner. It is unclear, however, what such useful inductive biases are. Initial approaches \citep{shah2021rrl,yuan2022pre, parisi2022unsurprising,gong2025autofocus} to this problem included simply reusing pretrained vision models trained to solve computer vision tasks like image recognition, zero-shot for control. Works like R3M \citep{nair2023r3m} and VIP \citep{ma2023vip} tried to utilize temporal consistency, enforcing images that are temporally close in a video demonstration are embedded close to each other. Other works like Voltron \cite{karamcheti2023voltron} and masked visual pretraining \cite{radosavovic2023real, xiao2022masked, seo2023masked, liang2024visarl} attempt to use image reconstruction as one such inductive bias.

\begin{figure*}
\begin{center}
\centerline{\includegraphics[width=.95\textwidth]{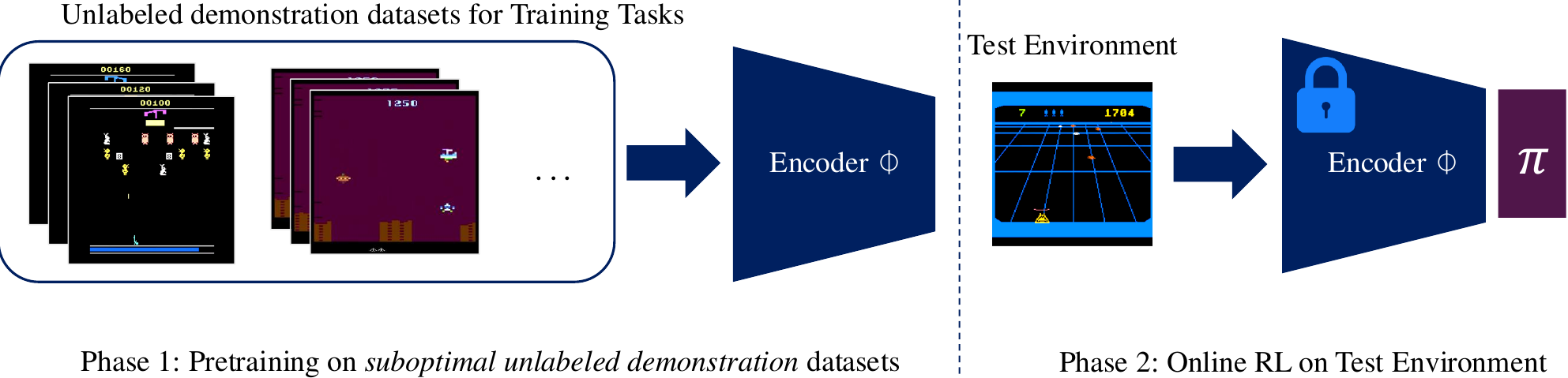}}
%\vspace{-8px}
\caption{\textbf{High-level overview of our setting.} The encoder $f_\phi$ is pretrained using suboptimal, unlabeled demonstration data from a set of train tasks, that is then reused for an unseen task. We evaluate pretrained encoders produced by our method and the baselines on the Atari and Navigation benchmarks.}
\label{intro}
\end{center}
\vskip -0.3in
\end{figure*}

While biases induced by pretraining objectives like image reconstruction and temporal consistency have been shown to greatly improve downstream policy performance, these pretraining objectives used to learn $f_\phi$ are \emph{distinct} from the downstream usage of $f_\phi$, e.g., the task of image reconstruction is very different from that of action prediction. There exists an unmet need for representation learning approaches that \emph{explicitly} encode information directly useful for downstream control during the process of learning $f_\phi$.

This is, of course, challenging---how do we encode control-specific information without actually training online on a control task? Our crucial insight is that encoding control-specific information in the representations generated by $f_\phi$ is possible by harnessing the power of Monte Carlo estimates of control heuristics computed offline using the suboptimal, unlabeled demonstration datasets.

%removed reward labels from the datasets.
Our key contribution is \emph{Value Explicit Pretraining (VEP)}, a contrastive learning approach that utilizes offline suboptimal, unlabeled demonstrations (without any action labels) to learn a representation for visual observations. Our method utilizes the insight that Monte Carlo value estimates across multiple tasks share a similar propensity of success, and, in tasks with related goals, also share a similar optimal policy. For example, in shooter games on Atari, despite differences in the visual appearances of adversaries, the strategy to effectively shoot them is similar. Our approach thus focuses on the similarity of progress towards the objective, as opposed to visual similarity.

VEP utilizes this intuition to learn an encoder using a contrastive loss which embeds observations with similar value function estimates across a set of training tasks near each other. We investigate the performance gains obtained by utilizing the VEP representation for policy learning, both on the training set of tasks and on visually distinct yet related held-out tasks. We experiment on the Atari benchmark and on a visual navigation benchmark comparing VEP to state-of-the-art methods like VIP \cite{ma2023vip} and SOM \cite{eysenbach2022contrastive}. We find up to a $2\times$ improvement in the rewards obtained on both benchmarks and $3\times$ improvement in sample efficiency of online RL algorithms trained on VEP.

\section{Related Work}

\textbf{Representation Learning for Robotics.} In addition to the general idea that representations have the role of encoding the essential information for a given task while discarding irrelevant aspects of the original data, typical state representation learning methods attempt to embed an observation into a latent representation that could be utilized by the downstream task \cite{DBLP:journals/nn/LesortRGF18}. It is also important that these methods produce a low dimensional representation that allows the control policy to \textit{efficiently} learn the downstream task. Traditionally, unsupervised methods like variational autoencoders \cite{DBLP:journals/corr/KingmaW13} learn disentangled representations that are used to correlate with underlying factors that cause variation in observation data \cite{higgins2017beta} for policy learning \cite{ha2018recurrent}. However, in many environments, these representations prove difficult to learn an optimal policy since the temporal structure is missing in these representations. \citet{anand2019unsupervised} explore this direction and learn representations by enforcing temporal structure through contrastive loss. However, they do not consider the generalization of the learned representations to unseen tasks.

National Eye

\noindent\textbf{Pretraining for RL.} Pretraining for representation learning, in the context of RL, involves learning transferable knowledge that helps the agent utilize its observations better \cite{xie2022pretraining}. Compared to traditional unsupervised methods for pretraining, the objective of self-supervised pretraining for RL is to learn representations by exploiting the underlying structure within the data distribution. Majority of the earlier \textit{online pretraining} works learn representations that model the task dynamics that is learned through expert videos during the RL procedure \cite{pathak2019self}. More recent \textit{offline pretraining} methods like \cite{schwarzer2021pretraining} build on the prior work \cite{anand2019unsupervised} by pretraining an encoder using unlabeled data and then finetune on a small amount of task-specific data. In comparison with these approaches, our method focuses on learning representations that not only aid in solving in-distribution tasks but also generalize to the out-of-distribution by learning representations that relate to general objectives and do not overfit to individual task-specific attributes.

\begin{figure*}[ht]
%\vskip 0.0in
\begin{center}
\centerline{\includegraphics[width=.95\textwidth]{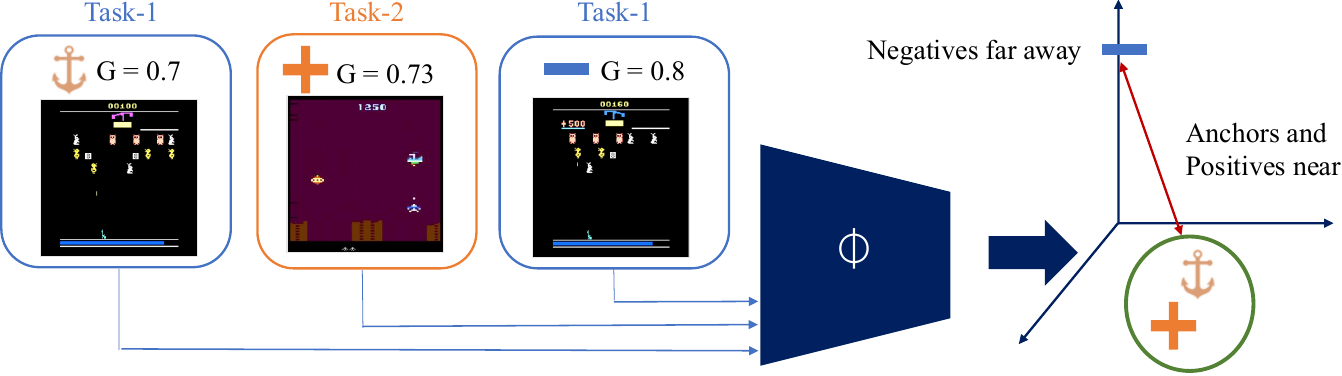}}
\caption{\textbf{Description of our method (VEP).} We compute Monte Carlo value estimates, as denoted by $G$, for each frame. We then use a contrastive learning-based pretraining method that learns task-agnostic representations based on $G$. The figure is a pictorial representation of a training scenario where we only have two tasks resulting in the \textbf{negative} getting sampled from the same task as the \textbf{anchor}. This may not be the case if we have more training tasks.}
\label{fig:approach}
\end{center}
\vskip -0.2in
\end{figure*}
%Another picture of a training instance where $b_T=4$ and $b_G=2$ is provided in the supplementary material.

%All the above methods perform pretraining as an optimization problem on a large set of trajectories. Our method, not only learns an encoder that performs the downstream task effectively but also embeds similar states of related tasks close to each other. This would enable the learned skills and policies from one task to be reused in another. 

\noindent\textbf{Transfer after Pretraining.} Transferring knowledge or skills learned from a given set of tasks to an unseen set of tasks is an active research area. Early works like progressive networks \cite{rusu2016progressive} attempt to solve it by reusing features learned from source tasks through adapters. \citet{gamrian2019transfer} perform image-to-image translation using GANs. However these methods are limited to predefined source or target domains. More recent works focus on the more challenging problem of using only expert videos for offline pretraining that could later be transferred to solve a novel downstream task. These methods have gained popularity in RL for their use of self-supervised pretraining based on \textit{contrastive learning} \cite{sermanet2018time}. Compared to these methods, our method only requires unlabeled data that consists of episodes that are not always successful in achieving the task objective. 

%left this just before deadline
%Recent works like \cite{eysenbach2022contrastive, ma2023vip} propose methods for learning an embedding space that could be used for zero-shot reward specification.

\noindent\textbf{Meta-learning for Transfer.} Several meta-learning approaches in robotics and RL share the paradigm of pretraining on training tasks and evaluating on test tasks. Meta-reinforcement learning operates through a two-phase structure where agents are meta-trained on a distribution of related tasks to learn a meta-policy that captures shared patterns, then evaluated on unseen test tasks with minimal additional training \cite{duan2016rl, finn2017model}. Information-theoretic task selection methods \cite{luna2020information} have also been developed to optimize which training tasks are used during meta-training to improve performance on test tasks. However, all these methods are limited due to their requirement of having labeled/demonstration datasets. Although value function based pretraining was studied in \cite{bhateja2023robotic,zhang2025rewind}, they also require multi-task demonstration data to learn representations unlike our method that relies only on unlabeled non-expert data.

\noindent\textbf{Baselines for VEP.} \textit{Value Implicit Pretraining (VIP)} \cite{ma2023vip} encodes the goal (positive) and the start (anchor) images close and the middle images (negatives) further away in the embedding space. By training on this objective through sampling multiple sub-episodes, the encoder recursively learns temporally smooth and continuous embeddings in a trajectory.
 \textit{Time Contrastive Learning (TCN)} involves sampling the positive within a certain margin distance $d_{thresh}$ from the anchor and a negative anywhere from the positive to the end of the trajectory \cite{sermanet2018time}. If the anchor is sampled at time instant $t_a$, positive is sampled at $t_p$ and the negative sampled at $t_n$, then $|t_n-t_a| > |t_p-t_a|$. It then uses the standard triplet loss for optimization, although other contrastive losses could also be used. Unlike TCN, \citet{eysenbach2022contrastive} sample the positives from \textit{State Occupancy Measure (SOM)} that could be embedded close to the anchor. The negative, on the other hand, is sampled anywhere from the other episodes of the same task or from the other tasks. State occupancy measure at a specific instant $t$ is a truncated geometric distribution $\textrm{Geo}_{t}^H (1 - \gamma)$  with probability mass re-distributed over the interval $[t, H]$, where $H$ is the horizon. \cite{mazoure2023accelerating}.

%\color{red}
%We also performed a comparison with an encoder pretrained by regressing the ground-truth value estimate which we call \textit{Value Regression (VR)}. We use a linear layer on top of the embedding dimension and train the entire model to predict the value estimate. After pretraining, we freeze the encoder and remove the linear layer that exposes the penultimate layer for Online RL training.
%\color{black}

\section{Problem Setting and Preliminaries}
Let $\mathcal{T}_{train} = \{\mathcal{T}_1, \mathcal{T}_2, ... \mathcal{T}_m\}$ be the set of training tasks with the associated suboptimal, unlabeled datasets that consist of a stream of images and sparse reward signals, denoted by $\mathcal{D}_{train}$. During pretraining, we assume that the encoder model $f_{\phi}$ parameterized by $\phi$ has access to $\mathcal{D}_{train}$. Data in $\mathcal{D}_{train}$ corresponding to $\mathcal{T}_i$ consists of a sequence of frames $\{o_t^i\}$ and sparse reward values $\{r_t^i\}$. The encoder $f_{\phi}$ learns to encode images/observations $o_t$ into an embedding $z_t$, which is taken as an input by the policy $\pi$ to perform a test-task. The set of test-tasks are denoted by $\mathcal{T}_{test} = \{\mathcal{T}_{m+1}, \mathcal{T}_{m+2}, ... \mathcal{T}_n\}$. Note that although $\mathcal{T}_{train} \cap \mathcal{T}_{test} = \varnothing$, all the tasks in $\mathcal{T}_{train} \cup \mathcal{T}_{test}$ share a semantically similar objective.

%Evaluating similarities across task objectives is an open problem in general \cite{DBLP:journals/pami/ZhuLJZ23}. Several methods have been proposed to quantify similarity across RL tasks \cite{DBLP:conf/icml/LazaricRB08}. This, however, is not our primary objective here. Instead..

For evaluation, we selected tasks that have semantically similar objectives, in three settings or benchmarks: 1) In Atari games, we obtain several shooter games that all contain the ``FIRE'' action in the action space and whose objective semantically relates to ``shoot up the enemy''. 2) For urban visual-based navigation, every task corresponds to navigating to the same goal destination with respect to the start location but in different cities; we use several cities and photographs taken along the available streets for the agent to navigate. 3) More challenging Ant locomotion benchmark \cite{ortiz2024dmc} consists of a quadruped Ant controlling its limbs to reach a goal location. The agent has access to different observations pertaining to a specific maze environment.

%All of them highly resemble the \texttt{Space-Invaders} concept, albeit with graphical and other variations: an army of alien enemies descends towards the bottom of the screen, where the agent's ship is, which can move left or right or shoot straight up.
%This, however, is not our primary objective here, and instead, we manually selected tasks that are intuitively similar, in two domains: 1) for urban visual-based navigation, we use maps of several cities and photographs taken along the available streets; to make tasks similar, we use a similar goal location relative to the agent's starting point, which eliminates the need to explicitly mention goal location. 2) In Atari games, we select the shooter games, which all highly resemble the \texttt{Space Invaders} concept, albeit with graphical and other variations: an army of alien enemies descends towards the bottom of the screen, where the agent's ship is, which can move left or right or shoot straight up.

For both our method and the baselines, the encoder $f_\phi$ is trained only using the unlabeled data $\mathcal{D}_{train}$ without any fine-tuning. The objective of our method is to efficiently learn the encoder using suboptimal, unlabeled data consisting of a sequence of observations and sparse reward signals, $\mathcal{D}_{train}$, from the source tasks $\mathcal{T}_{train}$, such that the embeddings from $f_{\phi}$ could be zero-shot transferred to unseen test tasks $\mathcal{T}_{test}$.

\subsection{Contrastive Representation Learning}
Typically, contrastive representation learning methods for RL utilize offline video demonstration datasets. These methods typically input a batch of anchors $\mathbf{o}_{an}$, positives $\mathbf{o}_{ps}$, and negatives $\mathbf{o}_{ng}$ and minimize a predetermined similarity metric that enables an encoder model to learn consistent and meaningful representations that can be used for downstream tasks. The earliest known formulation by \citet{schroff2015facenet} uses Euclidean distance to embed the positives and the anchor close to each other and the negatives far away from the anchor.
\begin{equation} \label{eq:triplet}
\mathcal{L}_{\textrm{triplet}} = \sum_{\mathbf{z} \in \mathcal{X}} \mathbf{max} \Big[\mathbf{0}, || \mathbf{z}_{an} - \mathbf{z}_{ps}||_2^2 - ||\mathbf{z}_{an} - \mathbf{z}_{ng} ||_2^2 + \epsilon \Big]
\end{equation}
%\normalsize

In the above equation $\mathbf{z}_{an}$, $\mathbf{z}_{ps}$ and $\mathbf{z}_{ng}$ represent the embeddings that are obtained after passing observations $\mathbf{o}_{an}$, $\mathbf{o}_{ps}$ and $\mathbf{o}_{ng}$ (anchors, positives and negatives) through the encoder network $f_{\phi}$. Other metrics like cosine similarity could also be used instead of Euclidean distance, to compute the similarity between embeddings. This loss is used by VEP and all our baselines except for VIP.

\begin{figure*}[ht]
\begin{center}\includegraphics[width=\textwidth]{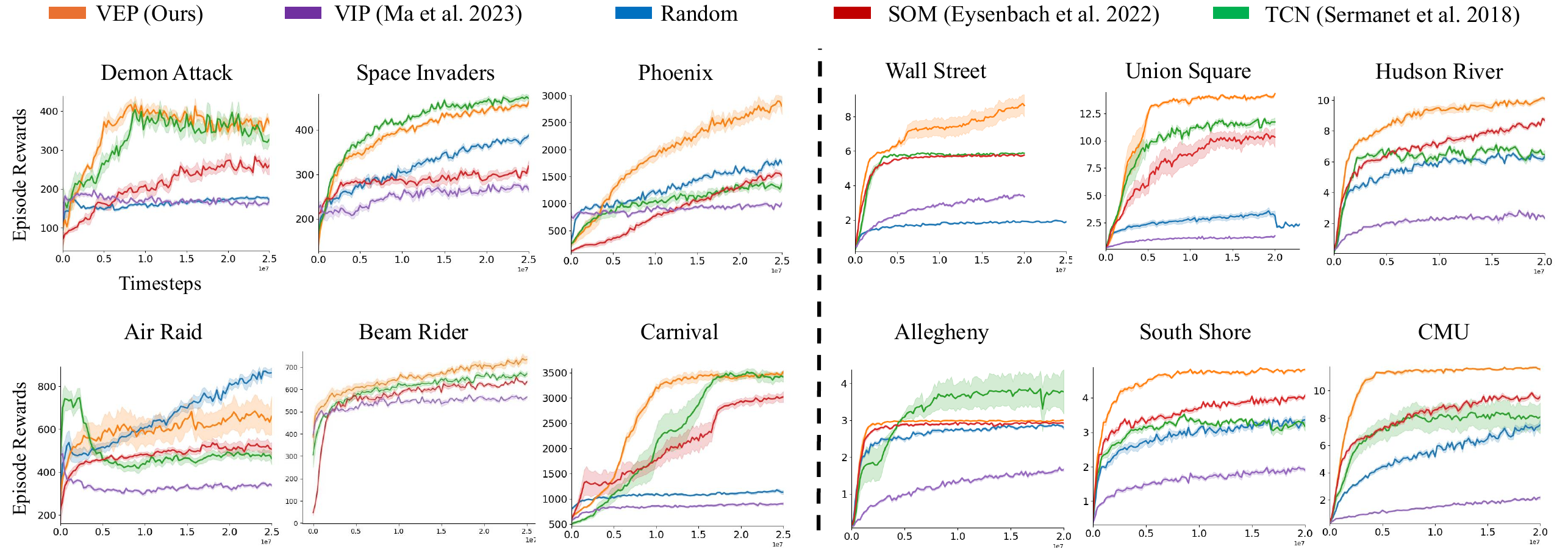}
\vskip -0.1in
\caption{(a). \textbf{Pretraining results on Atari (Left).} Performance of different pretraining methods on the respective games as mentioned above on the left. The encoder is pretrained only on the first two games (\texttt{Demon-Attack} and \texttt{Space-Invaders}) and is evaluated on the other out-of-distribution games. \textbf{Pretraining results on Navigation (Right).} Performance of different pretraining methods on the respective cities as mentioned above right. Similar to the Atari experiments, for all the baselines, suboptimal, unlabeled data from the first two tasks (\textit{Wall Street} and \textit{Union Square}) were used for pretraining. VEP representations improve PPO policy performance by up to $2\times$.}
\label{fig:mainexp}
\end{center}
\vskip -0.3in
\end{figure*}

Similar to recent methods like \cite{ma2023vip}, the InfoNCE \cite{oord2018representation} objective can also be used to optimize the encoder parameters. Unlike the triplet loss from Eq.~\eqref{eq:triplet}, InfoNCE permits utilizing \textit{multiple negative examples} for calculating the loss (via the expectation term in the denominator of Eq.~\eqref{eq:infonce}). As depicted below, InfoNCE aims to maximize mutual information of the anchors and positives. This loss is used by VIP:
\begin{equation}
\label{eq:infonce}
    \mathcal{L}_{\textrm{InfoNCE}} = \mathbb{E}_{\mathbf{z}_{ps}}\left[-\log \frac{\mathcal{S}_\phi \left(\mathbf{z}_{an}, \mathbf{z}_{ps}\right)}{\mathbb{E}_{\mathbf{z}_{ng}} \mathcal{S}_\phi\left(\mathbf{z}_{an}, \mathbf{z}_{ng}\right)}\right]
\end{equation}
where $\mathcal{S}_{\phi}$ computes the similarity between two embeddings in the $\phi$-representation space. In our experiments that use InfoNCE, we use cosine similarity for $\mathcal{S}_{\phi}$.

\subsection{Discounted Returns and Value Functions}
We consider a POMDP (Partially Observable Markov Decision Process) denoted by the tuple ($\mathcal{O}$, $\mathcal{S}$, $\mathcal{A}$, $p$, $\theta$, $r$, $T$, $\gamma$) representing an observation space $\mathcal{O}$, state space $\mathcal{S}$, action space $\mathcal{A}$, transition distribution $p$, emission function $\theta$, reward function $r$, time horizon $T$, and discount factor $\gamma$. An agent in state $\vb{s}_t$ takes an action $\vb{a}_t$ and consequently causes a transition in the environment through $p(\vb{s}_{t+1} \mid \vb{s}_t, \vb{a}_t)$. The agent receives the next observation $\vb{o}_{t+1}$ and reward $r_{t}$ that is calculated using the state $\vb{s}_t$ and action $\vb{a}_t$.
The objective for the agent is to learn a policy $\vb{\pi}$ which maximizes the expected discounted sum of rewards. The discounted sum of rewards at a state $\mathbf{s}_t$ in a trajectory $\tau$ is given by $\mathcal{G}$:
\begin{equation}
\mathcal{G}(\mathbf{s}_t, \tau) = r_{t}+\gamma r_{t+1}+\gamma^2 r_{t+2}+\cdots=\sum_{k=t}^T \gamma^{(k-t)} r_k
\label{eq:disc_return}
\end{equation}
%The expectation of this discounted return under the trajectory distribution $p(\tau)$ and the policy $\pi$ where $\tau$ is a trajectory of the form $(\mathbf{s}_t, \mathbf{a}_t, \mathbf{s}_{t+1}, \cdots)$ is often defined as the value of the state $\mathbf{s}_t$ under policy $\pi$, denoted by $\mathcal{V}^{\pi}(\mathbf{s}_t)$.
The expected return of state $\mathbf{s}_t$ under policy $\pi$ is called the value of $\mathbf{s}_t$ and denoted by $\mathcal{V}^{\pi}(\mathbf{s}_t)$.

\section{Method}
%\subsection{Pretraining Encoder}
%\subsection{Transfer using Contrastive learning}

The true value $\mathcal{V}^{\pi}(\mathbf{s}_t)$ of a state $\mathbf{s}_t$ under a policy $\pi$ intuitively defines the propensity for the success of solving a task by following policy $\pi$. If two states have similar value estimates, they likely have a similar expected return under $\pi$.

With this in mind, we now motivate VEP with an example. Consider the task of shooting an adversary in the Atari game of \texttt{Space-Invaders}. Assume that there exists an optimal policy for this task denoted by $\pi^*(\cdot \mid \mathbf{o}_t)$ which operates on image observations and the associated optimal value function $\mathcal{V}^{\pi^*}(\cdot)$. Now, consider a slightly perturbed version of this game in which all the adversaries are colored orange. If policy $\pi^*$ must solve this perturbed task, it must be invariant to the color of the adversary. One way to achieve this invariance is to enforce that the value estimates of states with similar propensities for success are similar, e.g., the value estimate of a state containing a bullet very close to an adversary should be the same regardless of whether the adversary is yellow or orange. VEP utilizes this very intuition by learning representations that induce such an invariance. We assume access to unlabeled data from suboptimal agents doing the task (playing the game), consisting of only observations and reward values (obtained sparsely) for the set of training tasks $\mathcal{T}_{train}$: This kind of data can be obtained from various online sources of gameplay, and does not contain any action labels. Further, it is assumed to be generated by a suboptimal agent that contains at least a few positive reward signals during gameplay. These suboptimal, unlabeled datasets consist of demonstration data that are \textit{not always guaranteed to succeed} in the task. If a sequence ends up with no reward at the terminal state, the last sub-sequence starting with the last non-zero reward and leading to the terminal state is rejected since all the frames in that sub-sequence end up with value estimates of $0$. Note that since the task objectives allow for sparse rewards during the task, we would still be using parts of the episode (sequence), although the episode might not achieve the final task completion.

\begin{figure*}[ht]
\begin{center}\includegraphics[width=\textwidth]{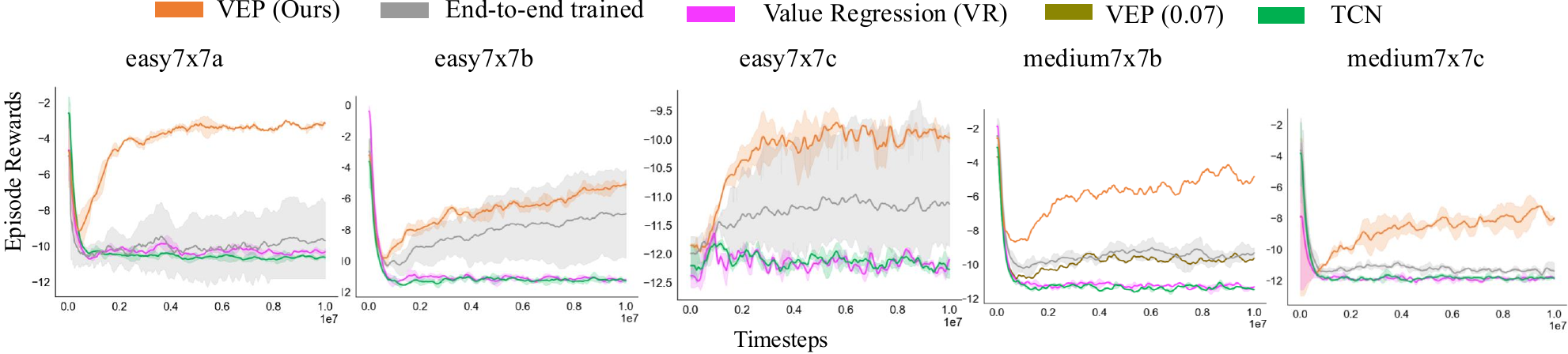}
\vspace{-20px}
\caption{\textbf{Pretraining results on Ant locomotion.} Performance of different pretraining methods on the respective mazes. The encoder is pretrained only on two mazes (\texttt{empty7x7} and \texttt{medium7x7a}) and is evaluated on the other out-of-distribution mazes that are shown above. The in-domain results are shown in Figure \ref{fig:antadd}. We also performed ablation studies by varying the $v_{thresh}$ parameter for VEP. Note that $v_{thresh} = 0.03$ for the optimal VEP baseline. We added another VEP baselines for \texttt{maze7x7b}, by varying the $v_{thresh}$ parameter, that are shown in the legend above as VEP ($v_{thresh}$). As the value threshold increases, the performance of our method degrades.}
\label{fig:antood}
\end{center}
\vspace{-20px}
\end{figure*}

We also do not have access to the true reward function, but operate under a sparse reward setting, assuming that a reward of $1$ at a few timesteps in the suboptimal, unlabeled data and $0$ everywhere else. We now compute a value estimate to each observation using Eq.~\eqref{eq:disc_return}. Ideally, this value estimate would be computed using $\mathcal{V}^{\pi^*}(\cdot)$, but since we do not have access to the true value function of the optimal policy, we utilize a Monte Carlo estimate of this using Eq.~\eqref{eq:disc_return}. The computation of value estimates is fully algorithmic and requires no human effort.

%\begin{figure*}[ht]
%\vskip -0.1in
%\begin{center}
%\begin{minipage}{0.3\textwidth}
%  \centering
%  \includegraphics[width=\textwidth]{pics/reward.png}
%\end{minipage}%
%\begin{minipage}{0.55\textwidth}
%  \centering
%  \includegraphics[width=.8\textwidth]{pics/VEP_pr.png}
%\end{minipage}
% \vskip -0.35in
%\caption{(a). \textbf{Reward functions for Navigation (left).} For a specific map, the agent spawns at a predetermined starting location (\textcolor{red}{red}), with the flexibility to initiate at a random location within a $r$-step to the fixed starting point. The sparsity of the rewards (\textcolor{brown}{brown} lines) that enable the agent to navigate to the goal (\textcolor{green}{green}) can be adjusted through the parameter $L$. (b). \textbf{Comparison with other existing pretrained models (right).} We show the bar plot that compares VEP with other existing pretrained models using the mean cumulative reward of %the policy on the out-of-distribution task. \color{red}(b). \textbf{Ant %Mujoco task} We show the bar plot that compares VEP with other existing %pretrained models using the mean cumulative reward of the policy on the %out-of-distribution task. **Erdem: Should I include a picture of Ant %maze task?**}\color{black}
%\label{fig:reward}
%\end{center}
%\vskip -0.1in
%\end{figure*}

%%Pls check!!
Having obtained several suboptimal, unlabeled datasets, for tasks in $\mathcal{T}_{train}$, and computed value estimates at each frame with $\mathcal{G}(\cdot)$ from Eq.~\eqref{eq:disc_return}, we now train the encoder $\phi$ using a contrastive learning objective. This procedure first involves sampling a scalar value estimate $g$ between $0$ and $1$ and then further sample multiple observations from $\mathcal{D}_{train}$ values within $v_{thresh}$ of $g$. Subsequently, an encoder $\phi$ is learned which embeds these observations close to each other. Consequently, observations with a similar propensity for success have similar embeddings. 

\subsection{Implementation}

To make the training computationally efficient, we preprocess $\mathcal{D}_{train}$ and save a dictionary that maps sorted Monte Carlo value estimates $\mathcal{G}(\cdot)$ to the indices of corresponding observations with the same Monte Carlo value estimate. This speeds-up the value lookup subroutines through binary search.

%perform experiments for value estimates.. and not distance measures

We first sample a batch of value estimates from the dataset determined by training batch size $b_{G}$. Next, we sample 3 training tasks ($\mathcal{T}_i$, $\mathcal{T}_j$ and $\mathcal{T}_k$) from $\mathcal{T}_{train}$ that we use to sample anchor, positive and negative respectively. Subsequently, the pretraining objective becomes:

\begin{align}
\max_{\phi}\! & \sum_{\mathcal{T}_i}\! \sum_{\mathcal{T}_j}\! \sum_{\mathcal{T}_k}\! \mathop{\mathbb{E}}_{g\sim \textrm{Unif}(\mathcal{G})} [\mathcal{S}_{\phi}(\mathbf{z}_{an}^{\mathcal{T}_i}, \mathbf{z}_{ps}^{\mathcal{T}_j}) - \mathcal{S}_{\phi}(\mathbf{z}_{an}^{\mathcal{T}_i}, \mathbf{z}_{ng}^{\mathcal{T}_k})]\nonumber
\label{eq:objective}
\end{align}
\normalsize
%\vspace{-.5pt}

where $\mathcal{G}\subset(0,1]$ is the set of Monte Carlo value estimates of all observations in $\mathcal{D}_{train}$. As mentioned before, $\mathcal{S}_{\phi}$ computes the similarity between 2 embeddings that are obtained from an encoder parameterized by weights $\phi$. $\mathbf{z}_{an}^{\mathcal{T}_i}$ corresponds to the embedding of the anchor, i.e., observation sampled from task $\mathcal{T}_i$ with a value estimate within an $v_{thresh}$ of $g$, and $\mathbf{z}_{ps}^{\mathcal{T}_j}$ corresponds to a positive, that is sampled from task $\mathcal{T}_j$, and has a value estimate within an $v_{thresh}$ of $g$ and the negative $\mathbf{z}_{ng}^{\mathcal{T}_k}$ is sampled from task $\mathcal{T}_k$ that has a value estimate beyond $v_{thresh}$ from $g$.

Intuitively, this objective encourages the positives and anchors from all the sampled tasks to embed near each other, using a value function estimate to organize the latent space of the learned encoder $\phi$.
For full implementation details like batch sizes etc., please refer to the website.

\begin{algorithm}
\footnotesize
\caption{Value Explicit Pretraining}
\begin{algorithmic}[1]
\REQUIRE $\mathcal{D}_{train}$, the set of suboptimal, unlabeled demonstrations collected from tasks $\{\mathcal{T}_i\}_{i=1}^{i=m}$
\REQUIRE Encoder $f_\phi$ parameterized by $\phi$
\REQUIRE $b_G$, as the batch size
\REQUIRE $v_{thresh}$ as the value thresholds
\REQUIRE $N$ as the number of iterations

\STATE Randomly initialize $\phi$
\STATE Compute value estimates $\mathcal{G}(.)$ for every frame $\mathbf{o_t}$ in $\mathcal{D}_{train}$
\STATE Remove all frames in the $\mathcal{D}_{train}$ having value estimate of 0
\STATE For every task $\mathcal{T}_i$, create a dictionary $\mathbf{V}^i$ mapping sorted value estimates as keys to list of frame indices in $\mathcal{D}_{train}$  
\WHILE{iterations until $N$}

\STATE Sample a $b_G$ sized batch of values $g \sim (0,1]$
\STATE For each sampled value $g$, sample a frame $o_{an} \sim \mathcal{T}_i$ that has a value estimate of within $v_{thresh}$ from $g$
\STATE Sample a positive $\mathbf{o}_{ps} \sim \mathcal{T}_j$ from $\mathcal{D}_{train}$ within $v_{thresh}$
\STATE Find negatives $\mathbf{o}_{ng} \sim \mathcal{T}_k$ that are beyond $v_{thresh}$ from $\mathbf{o}_{an}$

%\color{blue}
\STATE Estimate embeddings $\mathbf{z}_{an}$, $\mathbf{z}_{ps}$, $\mathbf{z}_{ng}$ for a batch of $\mathbf{o}_{an}$, $\mathbf{o}_{ps}$, $\mathbf{o}_{ng}$ by propagating through $f_{\phi}$
\STATE Compute contrastive loss using $\mathbf{z}_{an}$, $\mathbf{z}_{ps}$ and $\mathbf{z}_{ng}$
\STATE Optimize $\phi$

\ENDWHILE
\end{algorithmic}
\end{algorithm}

\section{Experimental Setup}
We study whether utilizing VEP as a pretraining objective to learn an encoder improves \textbf{(1)} policy learning on in-distribution tasks, i.e., those tasks for which data was available to pre-train the encoder and \textbf{(2)} whether the learned encoder aids transfer learning of new tasks. For additional details, please visit our website.
%We performed our experiments using the benchmark specified in the next paragraph. We used the RLLib library \cite{DBLP:conf/icml/LiangLNMFGGJS18} under the Ray ecosystem for all our RL experiments. We used PPO \cite{schulman2017proximal} for training the policy. For all the baselines, we use the same datasets as our method for pretraining. All the environments we use for evaluating the baselines and our method are long-horizon tasks with a horizon of 2000--4000 timesteps. Other additional details for our experimental setup are mentioned in the website.

% 2 games \textit{DemonAttack} and \textit{SpaceInvaders}
% The 5D action space consists of NoOp, Fire, Move Left, Move Right, Move Left and Fire, and Move Right and Fire.
% For example, the agent has access to shields in \textit{SpaceInvaders} that allow the agent to stay under to prevent getting shot.

\subsection{Environments}
\textbf{Atari.} We used six Atari games with ``FIRE'' in their action set, which all are \textit{Shoot'em up} games similar in spirit to Space Invaders. Although all the games share a common objective of shooting enemies that spawn from above, there are significant differences in appearances and dynamics across games. We then split these games into $\mathcal{T}_{\text{train}}$ and $\mathcal{T}_{\text{test}}$. For pretraining the encoder, we use suboptimal, unlabeled data, without action labels from the D4RL datasets \cite{fu2020d4rl}. The value estimates of each frame at timestep $t$ in a sequence are then computed using \eqref{eq:disc_return} with $T$ being the closest frame in the episode that obtains a reward.

%We truncate large sequences of raw videos into smaller videos by setting a state to terminal when the agent gets a reward. The terminal states across all the games are unified with the objective, as all the \textit{Shoot'em up} games deal with shooting the enemy objects.

%\begin{figure}[ht]
%\vskip -0.2in
%\begin{center}
%\centerline{\includegraphics[width=\columnwidth]{pics/reward.png}}
%\caption{\textbf{Reward functions for Navigation.} For a specific map, the agent %spawns at a predetermined starting location (red), with the flexibility to initiate at %a random location within a $r$-step to the fixed starting point. The sparsity of the %rewards (brown lines) that enable the agent to navigate to the goal (green) can be %adjusted through the parameter $L$.}
%\label{fig:rewad}
%\end{center}
%\vskip -0.5in
%\end{figure}

\begin{figure}[ht]
\vspace{-5px}
\begin{center}
\centerline{\includegraphics[width=\columnwidth]{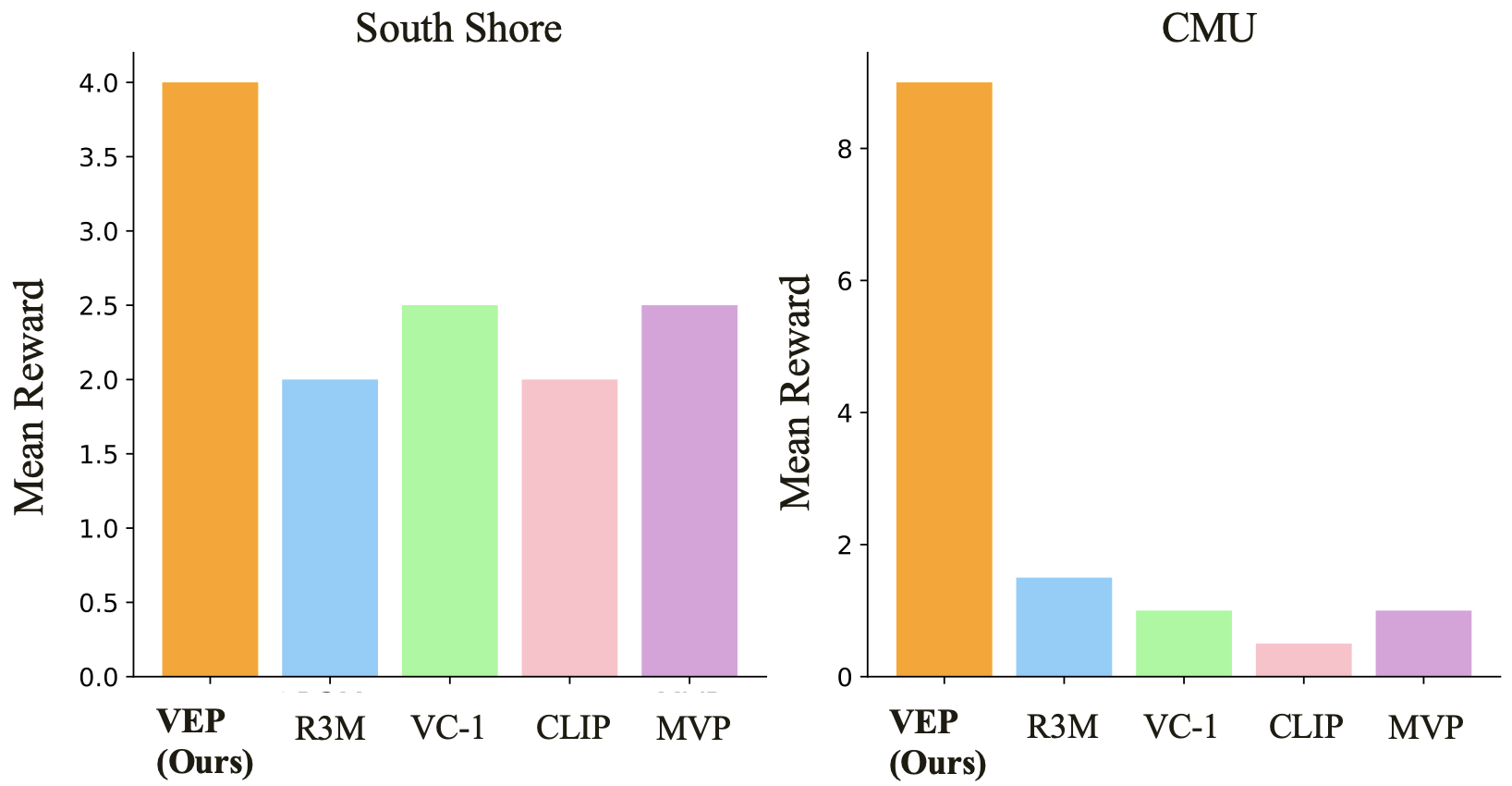}}
\vspace{-10px}
\caption{\textbf{Comparison with other existing pretrained models.} We show the bar plot that compares VEP with other existing pretrained models using the mean cumulative reward of the policy on the out-of-distribution task.}
\label{fig:prtr}
\end{center}
\vspace{-10px}
\end{figure}

\textbf{Navigation.} We build an engine that loads the StreetLearn dataset \cite{mirowski2019streetlearn} to perform visual navigation, based on gym \cite{brockman2016openai}. In a typical navigation task, the agent is designed to randomly respawn within a radius $r$ of a predetermined location $(src_x, src_y)$, with the objective reaching a goal location that is sampled within a radius $r$ of a location $(d_x + src_x, d_y + src_y)$. \textit{Sparse reward} acquisition is structured through a linear distribution of $L$ reward points (including the reward obtained upon reaching the target) uniformly spanning the starting point to the goal. The agent only earns \textit{sparse rewards} as it moves closer to the goal with each milestone radius from the goal destination. We have six cities for this benchmark, and we established consistent horizontal and vertical displacements $(d_x, d_y)$ between the starting and target points across all cities, avoiding the need for any explicit goal information (details are provided in the supplementary material). The agent is then expected to transfer to an unseen test city after learning from the suboptimal, unlabeled data obtained from a set of cities. Note that each task corresponding to a specific city is non-trivial since the agent needs to navigate in an unknown city with a different map and appearance. Tasks across all the cities are all solvable within a predefined horizon. To ensure that the dataset had at least a few sparse rewards, the start and the end locations of the path was chosen to be between the actual source and destination locations.

%Lastly, using a planner, we obtain the suboptimal, unlabeled data by randomly generating paths with a specific distance bound. For all the tasks, we set $L=15$ and $r=5$.

We use the same encoder architecture for both Atari and Navigation to embed pixel observation into vector space. To enable to temporal understanding of the state, all the embeddings in the past four timesteps are concatenated together and passed onto the policy. For Navigation, apart from the image embeddings, we also obtain odometry information $(odom_x, odom_y)$, of the agent, that is concatenated with the image embedding and passed into a linear layer. This enables the agent to understand its ego-centric pose with respect to the source location, which is crucial for understanding the objective and navigating to the goal.

\textbf{Ant locomotion in a maze.} To assess the efficacy of VEP on a more challenging benchmark, we use recently proposed Mujoco Ant based Visual Maze environment \cite{ortiz2024dmc}. This benchmark contains 7 different mazes and pictorially shown in Figure 4. \texttt{empty7x7}, \texttt{easy7x7} and \texttt{medium7x7} correspond to mazes having no obstacles, single obstacle and multiple obstacles respectively, and the alphabet at the end of each name corresponds to a specific maze. For pretraining, we used the suboptimal expert data from \texttt{empty7x7} and \texttt{medium7x7a}. The pretrain data for each maze consists of various visual variations of the maze, during evaluation there is a new variation that is not present in the offline pretrain dataset. This is important to note only for the in-domain evaluation (\texttt{empty7x7} and \texttt{medium7x7a}). For other out-of-domain mazes, we have both visual variations and structure variations pertaining to each maze. The pretrained data consists of the agent reaching a randomly sampled destination location from a randomly sampled source location. Unlike the data from the previous 2 benchmarks, there are no reward values associated with the pretrained data, and so we associated the reward $1$ at the last frame it reaches the destination, otherwise $0$ everywhere else.

We randomly sampled a source and destination position in each maze and made it fixed during the entire training process. The observation space is a 9 channel stacked image, i.e., RGB images corresponding to top, egocentric and following (visually captures ant joints) cameras. These are further stacked across three timesteps ($t$, $t-1$ and $t-2$). The agent obtains a 27 channel image and needs to output an 8 dimensional continuous control vector as an action. At every timestep, the agent obtains a reward equal to the change of distance to the goal. This is slightly different from the original reward function of providing the normalized negative distance to the goal at every timestep. The original reward function is primarily used as an evaluation metric for models trained using behavior cloning and did not show promising results when we used it to train our RL agent. We borrowed the architecture of the encoder from the original paper.

\begin{figure}[ht]
%\vskip -0.1in
\begin{center}
\centerline{\includegraphics[width=\columnwidth]{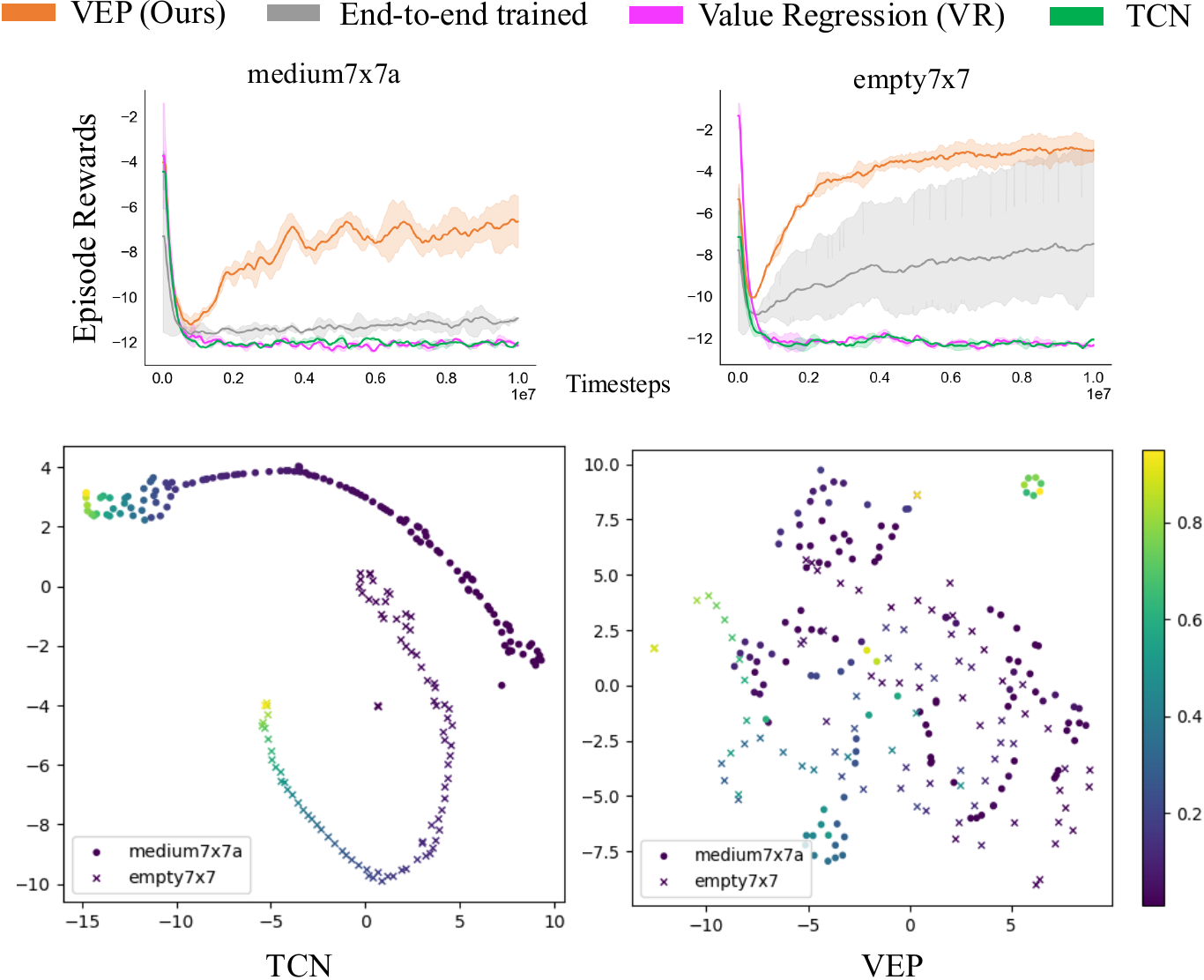}}
\vspace{-10px}
\caption{\textbf{Performance of our method on in-domain tasks in Ant locomotion benchmark (Top).} Top left and right corresponds to \texttt{medium7x7a} and \texttt{empty7x7} mazes.  \textbf{TSNE embeddings of our method compared to TCN (Bottom).} Color scale on the right corresponds to the value estimate. Notice how, even though the objective across both mazes is the same, TCN clusters the embeddings corresponding to the observations in the episodes for both the mazes separately, suggesting overfitting to the appearances of the maze. For the \underline{same episodes}, our method learns task-agnostic representations that help in efficient downstream policy learning.}
\label{fig:antadd}
\end{center}
\vskip -0.2in
\end{figure}

We first pretrain the encoder $f_{\phi}$ using the method described in the previous section and visually shown in Fig.~\ref{fig:approach} by using a sequence of unlabeled trajectories from both games. Once we obtain the pretrained encoder, we use an online RL algorithm, in our case PPO \cite{schulman2017proximal}, to train a policy.

\subsection{Results}
\textbf{Online RL experiments on Atari.} For experiments involving Atari games, we trained the policy by freezing the pretrained encoder, without any additional fine-tuning. The encoder is pretrained using offline suboptimal, unlabeled data from \texttt{Demon-Attack} and \texttt{Space-Invaders}, and evaluated on a set of in-distribution and out-of-distribution environments. We find that the pretrained encoder is able to outperform baselines on the in-distribution by nearly $25\%$. This margin is larger in the transfer experiments, most notably on \texttt{Phoenix}, with nearly $2\times$ improvement over baselines.  
% Although the main focus of our method is to evaluate a pretrained model on Out-of-domain environments, It's important to evaluate the model on in-domain environments, i.e. expert video data from those environments that is used for pretraining the encoder. As evident from many prior pretraining works, enabling a model to perform a task using the videos from the same task is still an ongoing research problem. The results are shown in Figure \ref{atari_prtr}.

\noindent\textbf{Online RL experiments on Navigation.} VEP outperforms all of our baselines by a larger margin in the navigation set as seen in Fig.~\ref{fig:mainexp}. VEP also outperformed the end-to-end trained baseline by achieving the same performance $2.1\times$ faster (results shown on the website). In addition, we evaluate our method on out-of-distribution tasks along with existing state-of-the-art \textit{Vision-language pretrained models}. Specifically, we compared our method (VEP) with CLIP \cite{radford2021learning}, MVP \cite{radosavovic2023real}, R3M \cite{nair2023r3m} and VC-1 \cite{majumdar2023we} and the results are shown in Figure~\ref{fig:prtr}. We hypothesize that the better performance of our method in the Navigation tasks was due to a more similar distribution of value estimates across the cities in the Navigation task, than the Atari games. Detailed specifications of the value estimates for all the Atari games and the cities in Navigation are described in the website.

\noindent\textbf{Online RL experiments on Ant locomotion.} Unlike single layer policies used in Atari and Navigation, we had to use a 2 layer policy with 256 hidden units that takes a 64 dimensional embedding and outputs an 8 dimensional action. The horizon of an episode was set to $1200$. Since continuous control tasks are more challenging, we had to increase the number of epochs per each update iteration from 10 to 20 compared to the discrete control tasks. As presented in Figure~\ref{fig:antood}, VEP outperforms all the baselines and achieves state of the art performance in both in-domain and out-of-domain mazes. For more details regarding the experiments on Ant locomotion, please visit our website.

% \textbf{Larger batch size and more iterations} All the baseline approaches we compared against had a fixed train batch size that is used for computing gradients. For VEP, we are required to use a larger have a train batch size $B_{T}$ (sampling a batch of specific value estimates) and the sample batch size $B_{S}$ (sampling a batch of data corresponding to each sampled value estimate). During optimization, the resultant batch is of size $B_{T}*B_{S}$ and is used to compute the gradients. To show that our method does not simply gain better performance because of the larger batch size, we compared a baseline approach (TCN) with batch size $B_{S}*B_{T}$ for $B_{S}*N$ iterations, where $B_{S}, B_{T}$ and $N$ is the sample batch size, train batch size and the number of iterations of VEP that is compared against TCN.

%\textbf{Early stopping to prevent overfitting.} For the Navigation task, we increased the number of training tasks from 2 to 4. We observed that the performance degraded in this setting when trained for an epoch. As shown in Figure~\ref{fig:add_exp}, when we reduce the number of iterations, the model retains the performance, which suggests that our method learns optimal representations with a fixed number of iterations irrespective of the dataset size. As the dataset size increases, early stopping can prevent overfitting.

\noindent\textbf{Ablation studies on VEP.} To understand the thresholding of the value estimates during sampling, we pretrained 3 different encoders using VEP with thresholds, $0.03, 0.07$ and $0.15$. We were unable to go lower than $0.03$, due to some samples not being able to qualify for any positive sample because of how long the episode was. We show the results in Fig.~\ref{fig:antood}. Though we show the results only for \texttt{medium7x7c} due to space constraints, we observed similar trends across all OOD mazes (additional plots available on our website).
%Compared to other baselines, since We also performed ablation studies on the sample batch size by increasing it to 2 and 4, corresponding to the increased number of positives per anchor, and the performance degraded. This, along with the degradation in performance with value regression baseline suggests that the improvement in performance comes from large sampling iterations combined with small batch size.

\noindent\textbf{Comparison with value regression baseline.}
Our method might seem similar to an objective that would learn representations by simply regressing the MC value estimate. We compared our method with an encoder pretrained by regressing the ground-truth value estimate which we call \textit{Value Regression (VR)}. We use a linear layer on top of the embedding dimension and train the entire model to predict the value estimate. After pretraining, we freeze the encoder and remove the linear layer that exposes the penultimate layer for Online RL training.

Based on the results comparing Value Regression baseline and the ablation studies performed, we can conclude that the strength of our method comes from large number of sampling the triplets (anchor, positive and negative) based on the value estimate. The contrastive learning loss ensures that the encoder learns representations useful for the downstream task. The value estimate is merely used as a tool for sampling the right triplets, and simply by regressing the encoder to predict the right value estimate does not help in learning useful representations compared to VEP, as evident in Figure. \ref{fig:antood}. Repeated sampling from a diverse set of tasks coupled with having anchor and positive within a range of value estimates helps the encoder learn task-agnostic representations.

\noindent\textbf{TSNE visualizations of embeddings.} For a randomly sampled episodes, each in the two in-domain tasks (\texttt{medium7x7a} and \texttt{empty7x7}), we plotted the TSNE visualizations of the embeddings obtained by passing the observations from these episodes through the pretrained encoders (Figure. \ref{fig:antadd}). Although embeddings from TCN are more temporally smooth, they nevertheless get clustered separately suggesting overfitting to the appearance of the environment. We argue that although temporal smoothness in the embedding space approximates an implicit value function \cite{ma2023vip} and is crucial for learning meaningful representations, this kind of task-specific clustering (Figure. \ref{fig:antadd}), assuming that these tasks share the same objective/goal, is detrimental for generalization for unseen tasks. Our method on the other hand learns both temporally consistant embeddings and task-agnostic representations. 

%Further, to ensure that gains demonstrated by VEP cannot be attributed only to larger batch size (Note that during each iteration, VEP uses a batch size of $b_{G}\times b_{T}$, whereas other baselines only use a batch of $b$), we doubled (since $b_T=2$) the batch size for TCN (denoted as TCN+), along with doubling the pretraining epochs, as seen in Figure~\ref{fig:add_exp}. The larger batch size for TCN still does not match the performance of VEP. 

\begin{figure}[ht]
\vspace{-5pt}
\begin{center}
\centerline{\includegraphics[width=\columnwidth]{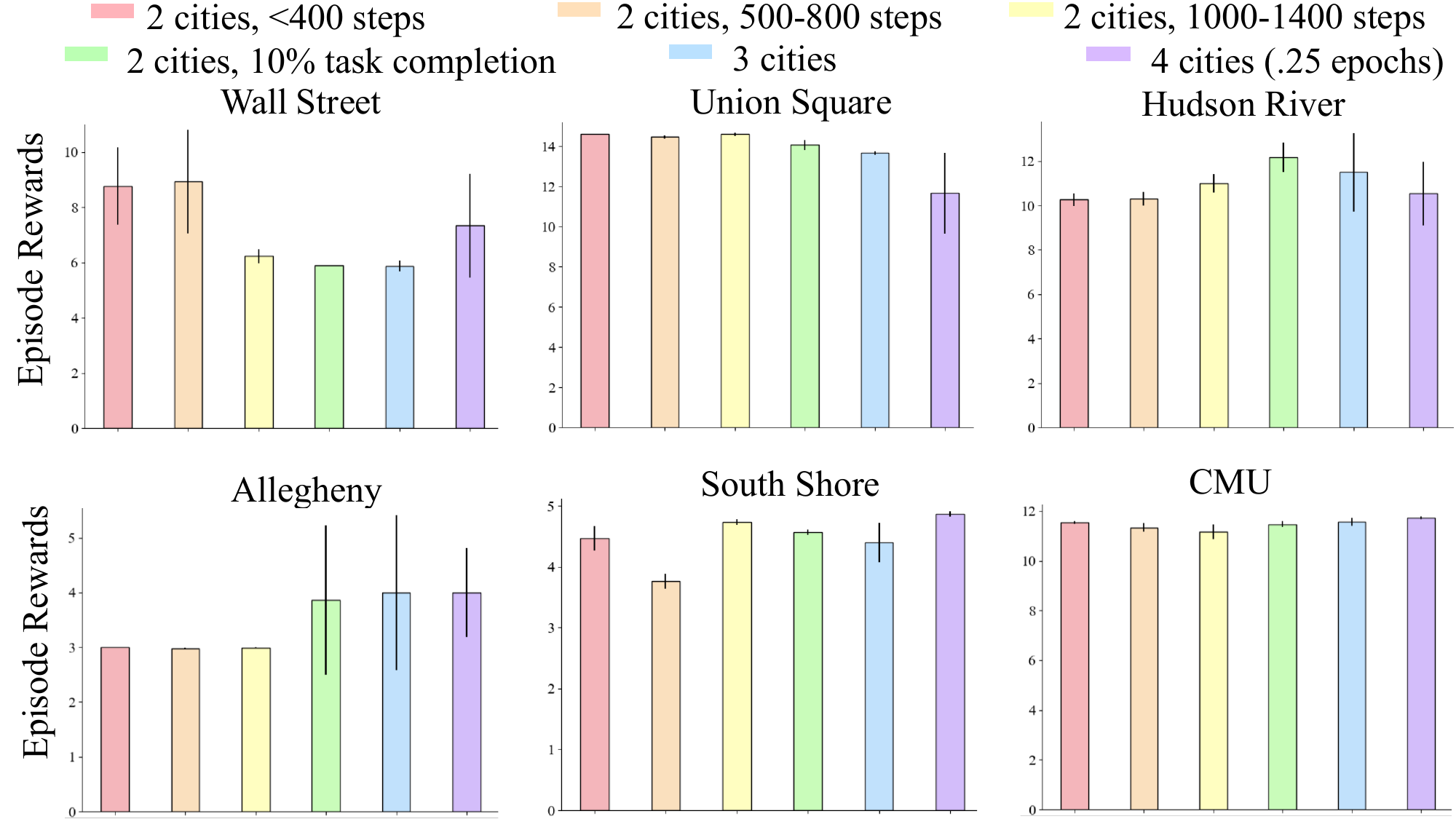}}
\vspace{-8px}
\caption{\textbf{Performance comparison on the quality of suboptimal, unlabeled data} Each of the above bar plots corresponds to the evaluation of the encoder in a different city. Each coloured bar corresponds to a specific suboptimal, unlabeled dataset used for pretraining. We also provide $95\%$ confidence intervals along with the mean cumulative reward.}
\label{fig:nav_qual}
\end{center}
\vskip -0.2in
\end{figure}

\noindent\textbf{Quality of suboptimal, unlabeled data.} We also evaluated our method by using different amounts of diversity and optimality in our dataset. Specifically, we compared with the datasets with episodes of length less than 400, 500-800, 1000-1400. All episodes in the respective datasets have a cumulative reward between 12-15. We also used suboptimal, unlabeled data that consist of episodes that complete $10\%$ of the actual task. Further, we also included episodes that consist of suboptimal data from 3 and 4 cities. As shown in Figure~\ref{fig:nav_qual}, we observe that our method is not affected by the suboptimality of the dataset. In other words, VEP uses non-zero value estimates during pretraining, and the suboptimality of the dataset has a negligible effect on the quality of representations learnt. Additionally, using multi-city datasets did help in learning better representations, except \texttt{Wall Street} and \texttt{Union Square}. In regards to those two cities, the original evaluations (red and orange bars) were obtained using a pretrained encoder trained on data from those very cities (In-domain), and hence performed the best.

\section{Conclusion}

We formulated a method to learn representations of states from different tasks solely based on the temporal distance to the goal frame. This way, the skills learned from the training tasks could be transferred to novel but related tasks. We showed the efficacy of our method by performing comprehensive evaluations on discrete and continuous control benchmarks like Atari, Visual Navigation and Ant locomotion.

For all the curent experiments, we used PPO for training the policy parameters. Future work would involve using more sample efficient algorithms like SAC \cite{haarnoja2018soft}. We also plan on experimenting with behavior cloning/few-shot imitation learning on manipulation benchmarks. Although our work shows the potential of intra-task pretraining, it is currently limited to the tasks that have similar objectives but still have different dynamics and appearances. We plan to study large-scale pretraining for more complex tasks in the future.

%\section*{Acknowledgments}
%This work was supported by the National Eye Institute (grant R61EY037527), the National Science Foundation (award 2318101), and a research grant from Sandia National Laboratories. The authors affirm that the views expressed herein are solely their own, and do not represent the views of the United States government or any agency thereof.
%\thanks{Manuscript
\renewcommand*{\bibfont}{\small}
\printbibliography

@article{xiao2022masked,
  title={Masked visual pre-training for motor control},
  author={Xiao, Tete and Radosavovic, Ilija and Darrell, Trevor and Malik, Jitendra},
  journal={arXiv preprint arXiv:2203.06173},
  year={2022}
}

@inproceedings{seo2023masked,
  title={Masked world models for visual control},
  author={Seo, Younggyo and Hafner, Danijar and Liu, Hao and Liu, Fangchen and James, Stephen and Lee, Kimin and Abbeel, Pieter},
  booktitle={Conference on Robot Learning},
  pages={1332--1344},
  year={2023},
  organization={PMLR}
}

@inproceedings{radosavovic2023real,
  title={Real-world robot learning with masked visual pre-training},
  author={Radosavovic, Ilija and Xiao, Tete and James, Stephen and Abbeel, Pieter and Malik, Jitendra and Darrell, Trevor},
  booktitle={Conference on Robot Learning},
  pages={416--426},
  year={2023},
  organization={PMLR}
}

@inproceedings{shah2021rrl,
  title={RRL: Resnet as representation for Reinforcement Learning},
  author={Shah, Rutav M and Kumar, Vikash},
  booktitle={International Conference on Machine Learning},
  pages={9465--9476},
  year={2021},
  organization={PMLR}
}

@inproceedings{parisi2022unsurprising,
  title={The unsurprising effectiveness of pre-trained vision models for control},
  author={Parisi, Simone and Rajeswaran, Aravind and Purushwalkam, Senthil and Gupta, Abhinav},
  booktitle={international conference on machine learning},
  pages={17359--17371},
  year={2022},
  organization={PMLR}
}

@article{yuan2022pre,
  title={Pre-trained image encoder for generalizable visual reinforcement learning},
  author={Yuan, Zhecheng and Xue, Zhengrong and Yuan, Bo and Wang, Xueqian and Wu, Yi and Gao, Yang and Xu, Huazhe},
  journal={Advances in Neural Information Processing Systems},
  volume={35},
  pages={13022--13037},
  year={2022}
}

@inproceedings{nair2023r3m,
  title={R3M: A Universal Visual Representation for Robot Manipulation},
  author={Nair, Suraj and Rajeswaran, Aravind and Kumar, Vikash and Finn, Chelsea and Gupta, Abhinav},
  booktitle={Conference on Robot Learning},
  pages={892--909},
  year={2023},
  organization={PMLR}
}

@inproceedings{ma2023vip,
title={{VIP}: Towards Universal Visual Reward and Representation via Value-Implicit Pre-Training},
author={Yecheng Jason Ma and Shagun Sodhani and Dinesh Jayaraman and Osbert Bastani and Vikash Kumar and Amy Zhang},
booktitle={The Eleventh International Conference on Learning Representations },
year={2023}
}

@inproceedings{karamcheti2023voltron,
 title={Language-Driven Representation Learning for Robotics},
 author={Karamcheti, Siddharth and Nair, Suraj and Chen, Annie S. and Kollar, Thomas and Finn, Chelsea and Sadigh, Dorsa and Liang, Percy},
 booktitle={Proceedings of Robotics: Science and Systems (RSS)},
 year={2023}
}

@article{eysenbach2022contrastive,
  title={Contrastive learning as goal-conditioned reinforcement learning},
  author={Eysenbach, Benjamin and Zhang, Tianjun and Levine, Sergey and Salakhutdinov, Russ R},
  journal={Advances in Neural Information Processing Systems},
  volume={35},
  pages={35603--35620},
  year={2022}
}

@inproceedings{DBLP:journals/corr/KingmaW13,
  author       = {Diederik P. Kingma and
                  Max Welling},
  title        = {Auto-Encoding Variational Bayes},
  booktitle    = {International Conference on Learning Representations (ICLR)},
  year         = {2014}
}

@inproceedings{bhateja2023robotic,
  title={Robotic Offline RL from Internet Videos via Value-Function Pre-Training},
  author={Bhateja, Chethan Anand and Guo, Derek and Ghosh, Dibya and Singh, Anikait and Tomar, Manan and Vuong, Quan and Chebotar, Yevgen and Levine, Sergey and Kumar, Aviral},
  booktitle={NeurIPS 2023 Foundation Models for Decision Making Workshop}
}

@inproceedings{haarnoja2018soft,
  title={Soft actor-critic: Off-policy maximum entropy deep reinforcement learning with a stochastic actor},
  author={Haarnoja, Tuomas and Zhou, Aurick and Abbeel, Pieter and Levine, Sergey},
  booktitle={International conference on machine learning},
  pages={1861--1870},
  year={2018},
  organization={Pmlr}
}

@inproceedings{zhang2025rewind,
 title={ReWiND: Language-Guided Rewards Teach Robot Policies without New Demonstrations},
 author={Zhang, Jiahui and Luo, Yusen and Anwar, Abrar and Sontakke, Sumedh A. and Lim, Joseph J. and Thomason, Jesse and Bıyık, Erdem and Zhang, Jesse},
 year={2025},
 booktitle={Conference on Robot Learning (CoRL)}
}

@article{gong2025autofocus,
  title={AutoFocus-IL: VLM-based Saliency Maps for Data-Efficient Visual Imitation Learning without Extra Human Annotations},
  author={Gong, Litian and Bahrani, Fatemeh and Zhou, Yutai and Banayeeanzade, Amin and Li, Jiachen and B{\i}y{\i}k, Erdem},
  journal={arXiv preprint arXiv:2511.18617},
  year={2025}
}

@inproceedings{liang2024visarl,
  title={Visarl: Visual reinforcement learning guided by human saliency},
  author={Liang, Anthony and Thomason, Jesse and B{\i}y{\i}k, Erdem},
  booktitle={2024 IEEE/RSJ International Conference on Intelligent Robots and Systems (IROS)},
  pages={2907--2912},
  year={2024},
  organization={IEEE}
}

@article{ortiz2024dmc,
  title={DMC-VB: A Benchmark for Representation Learning for Control with Visual Distractors},
  author={Ortiz, Joseph and Dedieu, Antoine and Lehrach, Wolfgang and Guntupalli, J Swaroop and Wendelken, Carter and Humayun, Ahmad and Swaminathan, Sivaramakrishnan and Zhou, Guangyao and L{\'a}zaro-Gredilla, Miguel and Murphy, Kevin P},
  journal={Advances in Neural Information Processing Systems},
  volume={37},
  pages={6574--6602},
  year={2024}
}

@article{luna2020information,
  title={Information-theoretic task selection for meta-reinforcement learning},
  author={Luna Gutierrez, Ricardo and Leonetti, Matteo},
  journal={Advances in Neural Information Processing Systems},
  volume={33},
  pages={20532--20542},
  year={2020}
}

@inproceedings{finn2017model,
  title={Model-agnostic meta-learning for fast adaptation of deep networks},
  author={Finn, Chelsea and Abbeel, Pieter and Levine, Sergey},
  booktitle={International conference on machine learning},
  pages={1126--1135},
  year={2017},
  organization={PMLR}
}

@article{duan2016rl,
  title={Rl$^2$: Fast reinforcement learning via slow reinforcement learning},
  author={Duan, Yan and Schulman, John and Chen, Xi and Bartlett, Peter L and Sutskever, Ilya and Abbeel, Pieter},
  journal={arXiv preprint arXiv:1611.02779},
  year={2016}
}

@inproceedings{sermanet2018time,
  title={Time-contrastive networks: Self-supervised learning from video},
  author={Sermanet, Pierre and Lynch, Corey and Chebotar, Yevgen and Hsu, Jasmine and Jang, Eric and Schaal, Stefan and Levine, Sergey and Brain, Google},
  booktitle={2018 IEEE international conference on robotics and automation (ICRA)},
  pages={1134--1141},
  year={2018},
  organization={IEEE}
}

@article{fu2020d4rl,
  title={D4rl: Datasets for deep data-driven reinforcement learning},
  author={Fu, Justin and Kumar, Aviral and Nachum, Ofir and Tucker, George and Levine, Sergey},
  journal={arXiv preprint arXiv:2004.07219},
  year={2020}
}

@article{mirowski2019streetlearn,
  title={The streetlearn environment and dataset},
  author={Mirowski, Piotr and Banki-Horvath, Andras and Anderson, Keith and Teplyashin, Denis and Hermann, Karl Moritz and Malinowski, Mateusz and Grimes, Matthew Koichi and Simonyan, Karen and Kavukcuoglu, Koray and Zisserman, Andrew and others},
  journal={arXiv preprint arXiv:1903.01292},
  year={2019}
}

@article{schulman2017proximal,
  title={Proximal policy optimization algorithms},
  author={Schulman, John and Wolski, Filip and Dhariwal, Prafulla and Radford, Alec and Klimov, Oleg},
  journal={arXiv preprint arXiv:1707.06347},
  year={2017}
}

@inproceedings{radford2021learning,
  title={Learning transferable visual models from natural language supervision},
  author={Radford, Alec and Kim, Jong Wook and Hallacy, Chris and Ramesh, Aditya and Goh, Gabriel and Agarwal, Sandhini and Sastry, Girish and Askell, Amanda and Mishkin, Pamela and Clark, Jack and others},
  booktitle={International conference on machine learning},
  pages={8748--8763},
  year={2021},
  organization={PmLR}
}

@article{majumdar2023we,
  title={Where are we in the search for an artificial visual cortex for embodied intelligence?},
  author={Majumdar, Arjun and Yadav, Karmesh and Arnaud, Sergio and Ma, Jason and Chen, Claire and Silwal, Sneha and Jain, Aryan and Berges, Vincent-Pierre and Wu, Tingfan and Vakil, Jay and others},
  journal={Advances in Neural Information Processing Systems},
  volume={36},
  pages={655--677},
  year={2023}
}

@article{mazoure2023accelerating,
  title={Accelerating exploration and representation learning with offline pre-training},
  author={Mazoure, Bogdan and Bruce, Jake and Precup, Doina and Fergus, Rob and Anand, Ankit},
  journal={arXiv preprint arXiv:2304.00046},
  year={2023}
}

@inproceedings{gamrian2019transfer,
  title={Transfer learning for related reinforcement learning tasks via image-to-image translation},
  author={Gamrian, Shani and Goldberg, Yoav},
  booktitle={International conference on machine learning},
  pages={2063--2072},
  year={2019},
  organization={PMLR}
}

@article{rusu2016progressive,
  title={Progressive neural networks},
  author={Rusu, Andrei A and Rabinowitz, Neil C and Desjardins, Guillaume and Soyer, Hubert and Kirkpatrick, James and Kavukcuoglu, Koray and Pascanu, Razvan and Hadsell, Raia},
  journal={arXiv preprint arXiv:1606.04671},
  year={2016}
}

@article{anand2019unsupervised,
  title={Unsupervised state representation learning in atari},
  author={Anand, Ankesh and Racah, Evan and Ozair, Sherjil and Bengio, Yoshua and C{\^o}t{\'e}, Marc-Alexandre and Hjelm, R Devon},
  journal={Advances in neural information processing systems},
  volume={32},
  year={2019}
}

@inproceedings{schroff2015facenet,
  title={Facenet: A unified embedding for face recognition and clustering},
  author={Schroff, Florian and Kalenichenko, Dmitry and Philbin, James},
  booktitle={Proceedings of the IEEE conference on computer vision and pattern recognition},
  pages={815--823},
  year={2015}
}

@inproceedings{higgins2017beta,
  title={beta-vae: Learning basic visual concepts with a constrained variational framework},
  author={Higgins, Irina and Matthey, Loic and Pal, Arka and Burgess, Christopher and Glorot, Xavier and Botvinick, Matthew and Mohamed, Shakir and Lerchner, Alexander},
  booktitle={International conference on learning representations},
  year={2017}
}

@article{ha2018recurrent,
  title={Recurrent world models facilitate policy evolution},
  author={Ha, David and Schmidhuber, J{\"u}rgen},
  journal={Advances in neural information processing systems},
  volume={31},
  year={2018}
}

@inproceedings{pathak2019self,
  title={Self-supervised exploration via disagreement},
  author={Pathak, Deepak and Gandhi, Dhiraj and Gupta, Abhinav},
  booktitle={International conference on machine learning},
  pages={5062--5071},
  year={2019},
  organization={PMLR}
}

@article{oord2018representation,
  title={Representation learning with contrastive predictive coding},
  author={Oord, Aaron van den and Li, Yazhe and Vinyals, Oriol},
  journal={arXiv preprint arXiv:1807.03748},
  year={2018}
}

@article{brockman2016openai,
  title={Openai gym},
  author={Brockman, Greg and Cheung, Vicki and Pettersson, Ludwig and Schneider, Jonas and Schulman, John and Tang, Jie and Zaremba, Wojciech},
  journal={arXiv preprint arXiv:1606.01540},
  year={2016}
}

@article{xie2022pretraining,
  title={Pretraining in deep reinforcement learning: A survey},
  author={Xie, Zhihui and Lin, Zichuan and Li, Junyou and Li, Shuai and Ye, Deheng},
  journal={arXiv preprint arXiv:2211.03959},
  year={2022}
}

@article{schwarzer2021pretraining,
  title={Pretraining representations for data-efficient reinforcement learning},
  author={Schwarzer, Max and Rajkumar, Nitarshan and Noukhovitch, Michael and Anand, Ankesh and Charlin, Laurent and Hjelm, R Devon and Bachman, Philip and Courville, Aaron C},
  journal={Advances in Neural Information Processing Systems},
  volume={34},
  pages={12686--12699},
  year={2021}
}

@article{DBLP:journals/nn/LesortRGF18,
  author       = {Timoth{\'{e}}e Lesort and
                  Natalia D{\'{\i}}az Rodr{\'{\i}}guez and
                  Jean{-}Fran{\c{c}}ois Goudou and
                  David Filliat},
  title        = {State representation learning for control: An overview},
  journal      = {Neural Networks},
  volume       = {108},
  pages        = {379--392},
  year         = {2018},
  url          = {https://doi.org/10.1016/j.neunet.2018.07.006},
  doi          = {10.1016/J.NEUNET.2018.07.006},
  bibsource    = {dblp computer science bibliography, https://dblp.org}
}

\end{document}